\pdfoutput=1

\documentclass[11pt]{article}

\PassOptionsToPackage{table}{xcolor}
\usepackage[preprint]{acl}
\usepackage{colortbl}

\usepackage{times}
\usepackage{latexsym}
\usepackage{amsfonts}
\usepackage{booktabs}
\usepackage{stfloats}
\usepackage[T1]{fontenc}

\usepackage[utf8]{inputenc}

\usepackage{microtype}
\usepackage{amsmath}

\usepackage{amssymb}
\usepackage{adjustbox}
\usepackage{algorithm}
\usepackage{algpseudocode}
\usepackage{inconsolata}

\usepackage{graphicx}

\title{\raisebox{-0.3\height}{\includegraphics[height=1.5em]{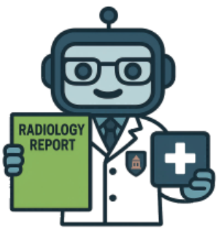}}~RadReason: Radiology Report Evaluation Metric with Reasons and Sub-Scores}

\author{
  \textbf{Yingshu Li\textsuperscript{1}},
  \textbf{Yunyi Liu\textsuperscript{1}},
  \textbf{Lingqiao Liu\textsuperscript{2}},
  \textbf{Lei Wang\textsuperscript{3}},
  \textbf{Luping Zhou\textsuperscript{1}\thanks{Corresponding author}}
\\
\\
  \textsuperscript{1}School of Electrical and Computer Engineering, University of Sydney, NSW 2006, Australia \\
  \textsuperscript{2}School of Computer Science, University of Adelaide, SA 5005, Australia \\
  \textsuperscript{3}School of Computing and Information Technology, University of Wollongong, NSW 2522, Australia
\\
}

\begin{document}
\maketitle
\begin{abstract}
Evaluating automatically generated radiology reports remains a fundamental challenge due to the lack of clinically grounded, interpretable, and fine-grained metrics. Existing methods either produce coarse overall scores or rely on opaque black-box models, limiting their usefulness in real-world clinical workflows. We introduce \textbf{RadReason}, a novel evaluation framework for radiology reports that not only outputs fine-grained sub-scores across six clinically defined error types, but also produces human-readable justifications that explain the rationale behind each score. Our method builds on Group Relative Policy Optimization and incorporates two key innovations: (1) \textbf{Sub-score Dynamic Weighting}, which adaptively prioritizes clinically challenging error types based on live F1 statistics; and (2) \textbf{Majority-Guided Advantage Scaling}, which adjusts policy gradient updates based on prompt difficulty derived from sub-score agreement. Together, these components enable more stable optimization and better alignment with expert clinical judgment. Experiments on the ReXVal benchmark show that RadReason surpasses all prior offline metrics and achieves parity with GPT-4-based evaluations, while remaining explainable, cost-efficient, and suitable for clinical deployment. Code will be released upon publication.
\end{abstract}

\section{Introduction}
\begin{figure}[t]
\centering
\centerline{\includegraphics[width=1\linewidth]{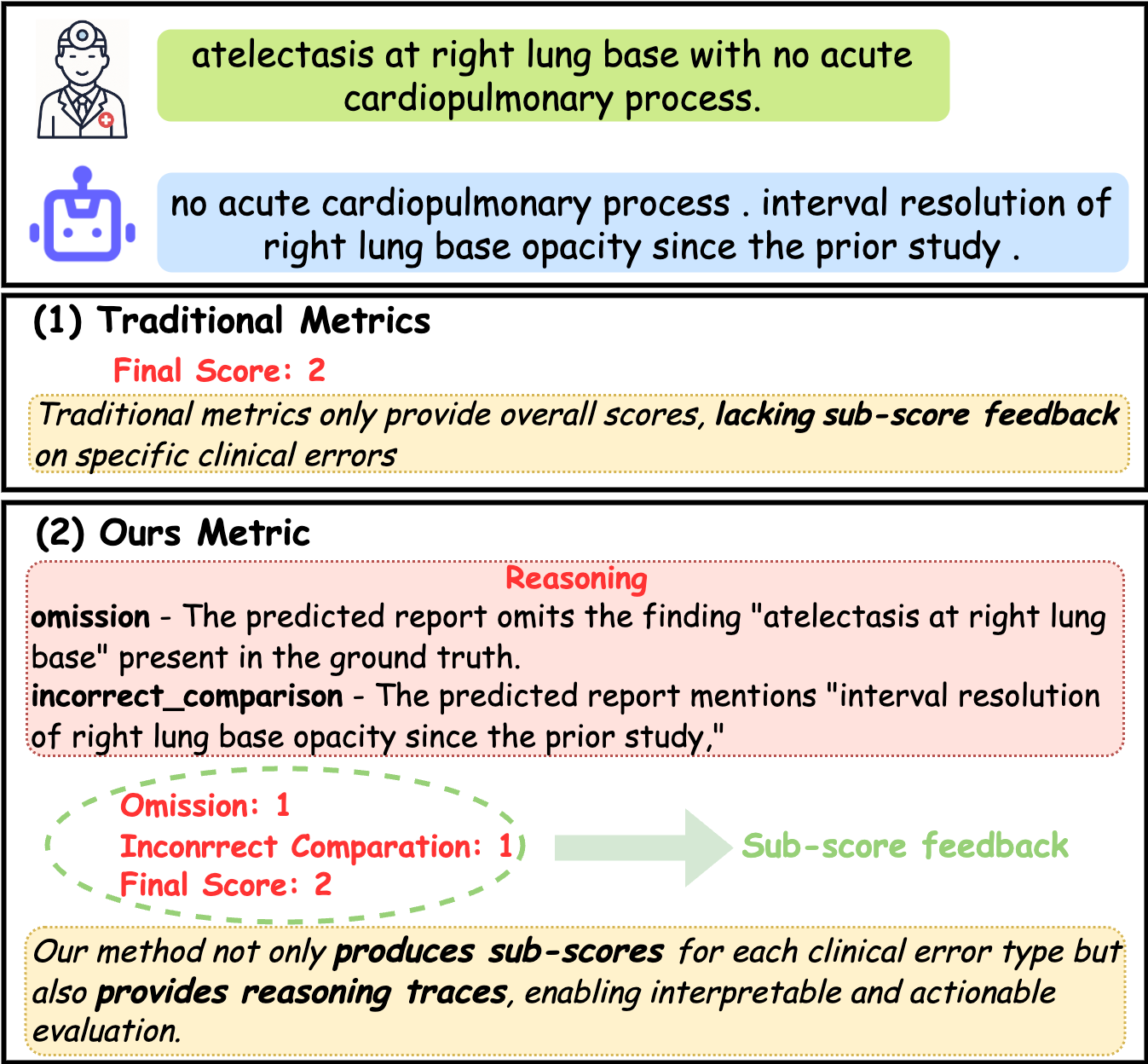}}
\caption{\textbf{Comparison of evaluation outputs across metric types.} Traditional metrics yield only overall scores, lacking insight into specific errors. Our method provides both sub-scores and reasoning traces, enabling interpretable and detailed evaluation.}
\vspace{-4mm}
\label{fig:brief}
\end{figure}

The automatic generation of radiology reports (RRG) from medical images has emerged as a pivotal task in clinical AI, offering the promise of reducing radiologists’ workload and improving diagnostic consistency~\cite{huang2023kiut,li2023comprehensive,li2024kargen,wang2024cxpmrg}. However, evaluating the quality of generated reports remains a fundamental challenge. Traditional natural language generation (NLG) metrics, such as BLEU, ROUGE, and METEOR~\cite{papineni2002bleu,banerjee2005meteor,lin2004rouge}, focus on word overlap and fail to capture clinically meaningful semantic differences, particularly in cases involving paraphrasing, negation, or subtle factual errors. Embedding-based metrics like BERTScore~\cite{zhang2019bertscore} improve semantic alignment but often overlook domain-specific entities critical to clinical interpretation. Structure-aware metrics, including RadGraph F1~\cite{jain2021radgraph} and CheXbert F1~\cite{smit2020chexbert}, incorporate medical knowledge but lack sensitivity to overall report quality and suffer from limited granularity. RadCliQ~\cite{yu2023evaluating} takes a step further by learning a regression model over multiple existing metrics to better approximate human judgments. More recently, large language models (LLMs) have been explored for radiology report evaluation~\cite{grattafiori2024llama,yang2024qwen2}. MRScore~\cite{liu2024mrscore} proposes a radiology-specific reward model to enable customized scoring, while Green~\cite{ostmeier2024green} evaluates factual correctness through explicit error-type matching, achieving close alignment with expert judgment. RaTEScore~\cite{zhao2024ratescore} improves semantic robustness using entity-aware similarity that handles synonyms and negations. In online GPT-based applications, CheXprompt~\cite{zambrano2025clinically} and FineRadScore~\cite{huang2024fineradscore} leverage GPT-4 to identify clinical error types and generate detailed corrections through few-shot prompting.

Despite recent progress, existing evaluation methods, summarized in Figure~\ref{fig:brief}, face two major limitations: (1) most systems produce only a single overall score, lacking error-type granularity; and (2) few provide explicit reasoning for \textit{why} a particular score was assigned, limiting clinical usability and model transparency. To address these issues, we introduce RadReason, a novel evaluation framework that decomposes report quality into six clinically defined error dimensions (e.g., false prediction, omission, incorrect location)~\cite{yu2023evaluating}, and produces both structured sub-scores and corresponding natural language explanations for each generated report. For example, “the report failed to mention left-sided effusion $\rightarrow$ omission errors = 1”. Technically, RadReason is trained via Group Relative Policy Optimization (GRPO)~\cite{shao2024deepseekmath,guo2025deepseek}, a reinforcement learning paradigm that models preferences over grouped completions. However, unlike prior work that yields a single scalar score, our framework predicts six distinct sub-scores, each corresponding to a specific error type. This setting introduces two primary challenges: (1) some error types are rare or harder to handle, requiring adaptive prioritization; and (2) report prompts vary in difficulty, with some inducing consistent completions while others exhibit high disagreement. To mitigate these challenges, we incorporate two auxiliary mechanisms: (1) Sub-score Dynamic Weighting, which dynamically adjusts reward weights according to the F1 performance of each error type to target areas of weakness; and (2) Majority-Guided Advantage Scaling, which leverages majority vote statistics to estimate sample difficulty and scale policy gradient updates accordingly. Experiments on the ReXVal benchmark~\cite{yu2023radiology} demonstrate that RadReason achieves state-of-the-art correlation with expert ratings, outperforming all prior offline metrics while remaining interpretable. Our key contributions include:

\noindent(1) We introduce \textbf{RadReason}, one reward-optimization-based evaluation framework for radiology report generation that outputs structured sub-scores and natural language explanations.

\noindent(2) We propose two novel training strategies, Sub-score Dynamic Weighting and Majority-Guided Advantage Scaling, to enhance clinical sensitivity.

\noindent(3) RadReason achieves state-of-the-art human alignment on ReXVal, while remaining efficient, interpretable, and extensible to new evaluation criteria.

\section{Related Works}
\subsection{Evaluation Metrics for Radiology Reports.}
Evaluating radiology report generation (RRG) requires metrics that assess both linguistic fluency and clinical correctness~\citep{yu2023evaluating}. Traditional NLG metrics—such as BLEU~\citep{papineni2002bleu}, ROUGE~\citep{lin2004rouge}, and METEOR~\citep{banerjee2005meteor}, focus on surface-level overlap, failing to capture paraphrasing, negation, or subtle clinical inaccuracies. As a result, semantically accurate reports with alternative phrasing may be penalized unfairly.
To address semantic fidelity and clinical content, several domain-aware metrics have been proposed. CheXbert F1~\citep{smit2020chexbert} leverages a pathology classification model trained on 14 thoracic diseases, while RadGraph F1~\citep{jain2021radgraph} evaluates factual correctness via structured entity-relation graphs. RadCliQ~\citep{yu2023evaluating} combines multiple such metrics using a regression model aligned with human annotations, improving correlation but offering limited interpretability.
Recent work leverages large language models (LLMs) for radiology report evaluation~\citep{grattafiori2024llama,yang2024qwen2}. MRScore~\citep{liu2024mrscore} trains a radiology-specific reward model to define a custom scoring framework. Green~\citep{ostmeier2024green} evaluates factual correctness and clinical significance based on matched findings and identified errors, demonstrating strong alignment with expert assessments. RaTEScore~\citep{zhao2024ratescore} is an entity-aware metric that handles synonyms and negations robustly, further aligning with human judgments. 
Several methods have also explored prompting commercial LLMs such as GPT-4. CheXprompt~\citep{zambrano2025clinically} use GPT-4~\citep{achiam2023gpt} to detect six specific error types: false positives, omissions, incorrect location, incorrect severity, irrelevant comparisons, and missing comparative statements. FineRadScore~\citep{huang2024fineradscore} applies few-shot prompting to elicit line-by-line corrections and clinical severity ratings for each identified error. However, these methods raise privacy concerns and depend on online access, which limits their practical deployment. Moreover, a common gap remains: most metrics provide only a single overall score, lacking fine-grained sub-aspect feedback or a clear rationale behind score assignments. To address this, we develop an offline evaluation framework capable of producing clinically aligned sub-scores and reasoning explanations per error type, advancing both interpretability and practical applicability.

\subsection{Reasoning in Large Language Models.}
Recent advances in large language models (LLMs) have shown strong capabilities in mimicking human-like reasoning, particularly by decomposing complex tasks into structured intermediate steps. This paradigm—often referred to as explicit reasoning—enables models to engage in interpretable, step-by-step thinking before arriving at a final output. A variety of techniques have been proposed to follow this approach, including prompting-based strategies such as Chain-of-Thought (CoT)~\citep{wei2022chain}, planning-oriented methods like Graph-of-Thought and Tree-of-Thought~\citep{besta2024graph,yao2023tree}. Beyond prompting, supervised fine-tuning (SFT) on datasets annotated with reasoning traces~\citep{kumar2025llm} can further enhance reasoning ability, but requires high-quality, labor-intensive annotations that limit scalability. To overcome this, recent work has adopted reinforcement learning (RL) to induce reasoning behaviors without explicit supervision. For example, DeepSeek-R1~\citep{guo2025deepseek} introduces an RL framework where the model is guided to generate reasoning trajectories followed by answers, and is rewarded based on final correctness, enabling learning from answer-only datasets. Building on this idea, we adopt an RL framework tailored to the evaluation metric. Our goal is to teach the model to both assign fine-grained sub-scores and generate natural language justifications. This enables interpretable, aspect-specific evaluation, critical for clinical auditability and decision support.

\section{Methods}
\begin{figure*}[ht]
\centering
\centerline{\includegraphics[width=1\linewidth]{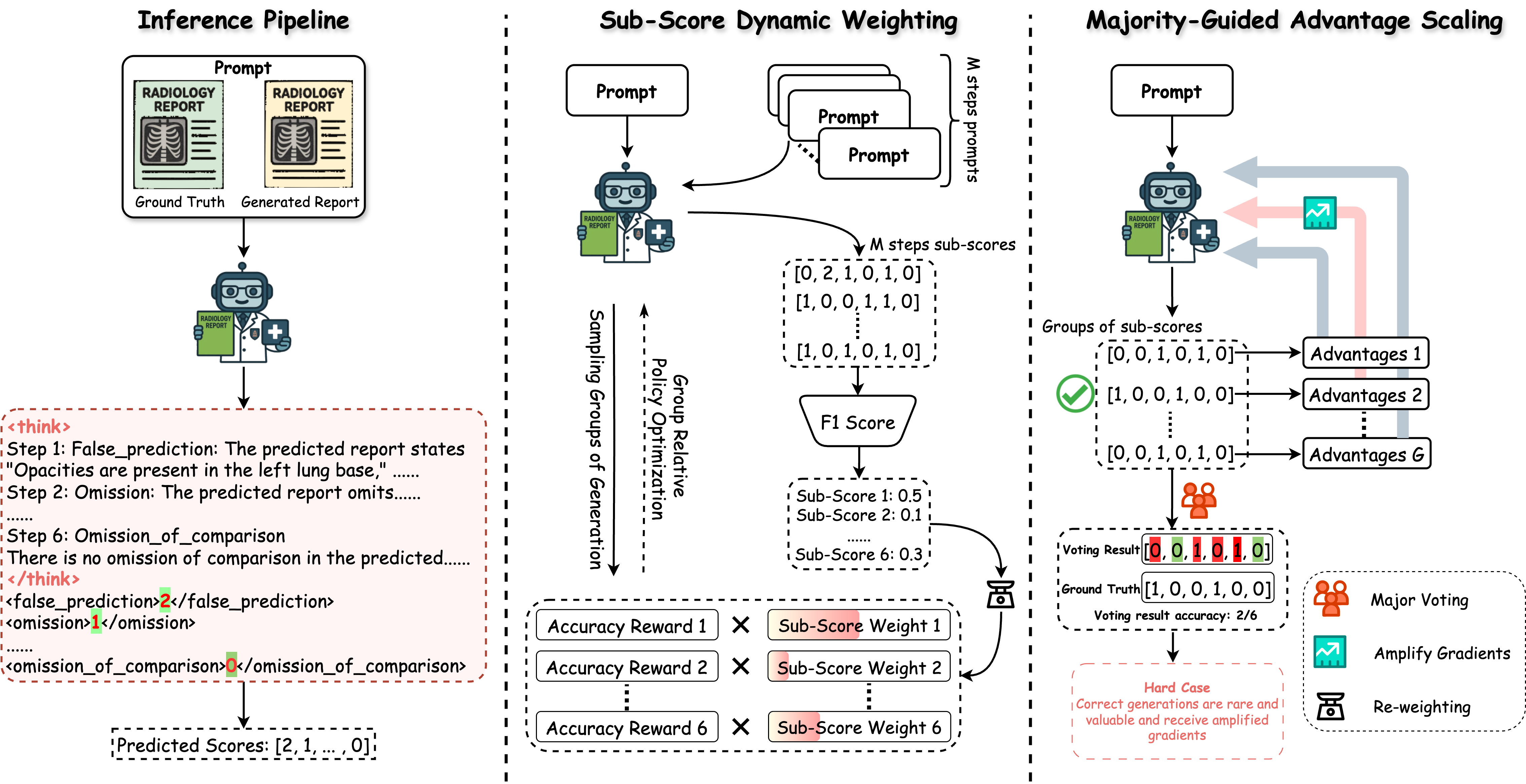}}
\caption{Overview of our training framework.
\textbf{Left:} The model produces detailed sub-scores and explanations across six clinically defined error types.
\textbf{Middle:} During training, we apply Dynamic Sub-score Weighting by periodically computing F1 gaps and reweighting each dimension’s reward accordingly.
\textbf{Right:} We introduce Majority-Guided Advantage Scaling, which estimates prompt difficulty by comparing majority-voted sub-scores across completions with the ground truth. Our method amplifies gradients for correct completions on hard prompts, while penalizing incorrect completions more heavily on easy ones.}
\label{fig:overall}
\end{figure*}

We propose a reinforcement learning framework for radiology report evaluation that produces interpretable sub-scores and explanations across clinical error types, as shown in Figure~\ref{fig:overall}. Our method integrates three key components: (1) sub-score prediction with natural language reasoning; (2) \textbf{Dynamic Sub-score Weighting}, which emphasizes clinically challenging aspects by adapting to per-dimension performance; and (3) \textbf{Majority-Guided Advantage Scaling}, which modulates policy updates based on prompt difficulty to reward rare but valuable completions.

\subsection{Background}
Group Relative Policy Optimization (GRPO)~\citep{guo2025deepseek} is a reinforcement learning algorithm designed for optimizing language models using group-wise preference signals. Unlike pairwise methods like DPO~\citep{rafailov2023direct}, GRPO compares multiple completions within a group and computes relative advantages to guide policy updates. For each prompt, GRPO samples a set of completions and assigns rewards via a reward model. Let $\mathbf{r} = \{r_1, r_2, \cdots, r_G\}$ denote the set of rewards for a group of $G$ completions for question $q$. The advantage of each completion is normalized within its group:
\begin{equation}
\hat{A}_{i,t} = \tilde{r}_i = \frac{r_i - \text{mean}(\mathbf{r})}{\text{std}(\mathbf{r})}.
\end{equation}
The GRPO loss function can be defined as:
\begin{adjustbox}{width=\linewidth}
\begin{minipage}{\linewidth}
\small
\begin{align}
\mathcal{L}_{\text{GRPO}}(\theta) = 
- \frac{1}{G} \sum_{i=1}^{G} \sum_{t=1}^{|o_i|} \Bigg[
\frac{\pi_\theta(o_{i,t} \mid q, o_{i,<t})}
     {\pi_{\theta_{\text{old}}}(o_{i,t} \mid q, o_{i,<t})}
\hat{A}_{i,t} \notag \\
\quad - \beta \, D_{\text{KL}} \left( \pi_\theta \,\|\, \pi_{\theta_{\text{ref}}} \right)
\Bigg], 
\end{align}
\end{minipage}
\end{adjustbox}
where \( o_i \) is the set of sampled completions for the \( i \)-th prompt, and  \( \pi_\theta(o_{i,t} \mid q, o_{i,<t}) \) denotes the token-level likelihood under the current policy.

\subsection{Rewards Construction}
To guide GRPO training, we design three reward functions that capture distinct aspects of evaluation quality: reasoning completeness, formatting correctness, and sub-score accuracy. For each sampled completion, we compute rewards as follows:

\noindent\textbf{Structured Reasoning Reward.}~~This reward promotes interpretability by verifying whether the model explicitly discusses all six clinical error types proposed in~\cite{yu2023evaluating}: false prediction, omission, incorrect location, incorrect severity, incorrect comparison, and omission of comparison. We use regular expressions to detect whether each aspect is addressed with appropriate structural cues (e.g., “Step 1: false prediction”).

\noindent\textbf{Format Reward.}~~We enforce output by requiring a single <think>...</think> block containing the reasoning text, and exactly one valid numerical value within each sub-score tag (e.g., <omission>1.0</omission>). This promotes format consistency and ensures reliable sub-score extraction.

\noindent\textbf{Accuracy Reward.}~~This component measures how closely the predicted sub-scores align with ground-truth annotations. For each generated report, we extract predicted sub-scores via regular expressions and compare them against the corresponding ground-truth values. Unlike previous GRPO-based approaches that use binary (0/1) reward signals, penalizing any deviation from the ground truth with a zero reward, such rigid schemes fail to differentiate between near-correct and completely incorrect predictions, leading to sparse and unstable learning signals. To address this, we introduce a smooth gaussian reward function that penalizes prediction errors based on the squared distance from the ground truth, thereby providing a more stable and informative learning signal. The reward for sub-score $j$ is computed as:
\begin{equation}
r^{(j)} =  \exp\left( -\frac{(\text{pred}^{(j)} - \text{gt}^{(j)})^2}{2\sigma^2} \right),
\end{equation}
with \( \sigma = 0.5 \) as the standard deviation controlling the tolerance to prediction error.

The sub-score rewards are then averaged across all \( K = 6 \) dimensions:
\begin{equation}
r_{\text{sub}} = \frac{1}{K} \sum_{j=1}^{K} r^{(j)}.
\end{equation}

In addition to evaluating each sub-score individually, we incorporate a total-score alignment term that encourages the sum of predicted sub-scores to match the ground-truth sum:
\begin{equation}
r_{\text{total}} = \exp\left( -\frac{(\hat{s}_{\text{total}} - s_{\text{total}})^2}{2\sigma^2} \right),
\end{equation}
where \( \hat{s}_{\text{total}} \) and \( s_{\text{total}} \) denote the predicted and ground-truth total scores respectively.

The final accuracy reward is a weighted combination of the two components:
\begin{equation}
r_{\text{acc}} = r_{\text{sub}} + r_{\text{total}}.
\end{equation}
The final reward for one output is:
\begin{equation}
r = r_{reasoning} + r_{format} + r_{acc}
\end{equation}
\subsection{Sub-score Dynamic Weighting}
While the accuracy reward evaluates each sub-score independently, it implicitly assumes equal importance across all error types. In practice, these aspects differ in frequency, ambiguity, and clinical impact. For example, \textit{omission} and \textit{false prediction} errors occur more frequently but tend to be ambiguous, while incorrect comparison is rarer yet often carries critical clinical implications. Consequently, uniformly averaging sub-score rewards can lead to biased optimization, with frequent and easier-to-learn aspects disproportionately influencing the gradient updates.

To address this imbalance, we propose a Sub-score Dynamic Weighting \textbf{(SDW)} strategy, which dynamically emphasizes sub-scores where the model exhibits weaker performance. Concretely, every $M$ steps, we compute the F1 score \( F1^{(j)} \) for each aspect \( j \in \{1, ..., K\} \), and define the relative performance gap as: $\Delta_j = \bar{F}1 - F1^{(j)}, \quad \text{where} \quad \bar{F}1 = \frac{1}{K} \sum_{j=1}^K F1^{(j)}$.

We then convert these difficulty scores into a normalized set of weights using a softmax function:
\begin{equation}
w_j = 1 + \frac{\exp(\alpha \cdot \Delta_j)}{\sum_{k=1}^K \exp(\alpha \cdot \Delta_k)},
\end{equation}
where \( \alpha \) is a temperature hyperparameter controlling the sharpness of the focus on under-performing dimensions.

We then compute the final sub-score reward as a weighted average across dimensions:
\begin{equation}
r_{\text{sub}}^{\text{dyn}} = \frac{1}{K}\sum_{j=1}^{K} w_j \cdot r^{(j)}.   
\end{equation}

This strategy allows the model to continuously reallocate supervision toward clinically challenging or underperforming aspects. Unlike static reward averaging, SDW encourages more balanced learning and improves robustness across diverse error types.

\begin{algorithm}[h]
\small
\caption{\small Sub-score Dynamic Weighting}
\label{alg:subscore_only}
\begin{algorithmic}[0]  
\Require F1 scores, update interval $M$, temperature $\alpha$
\State Initialize $w_j \gets 1$ for $j = 1,\dots,K$
\For{each training step $t = 1$ to $T$}
    \If{$t \mod M = 0$}
        \State Calculate $F1^{(j)}$ score for each aspect $j$
        \State Compute average F1: $\bar{F}1 \gets \frac{1}{K} \sum_{j=1}^{K} F1^{(j)}$
        \For{each aspect $j = 1$ to $K$}
            \State Compute F1 gap: $\Delta_j \gets \bar{F}1 - F1^{(j)}$
        \EndFor
        \State Update weights with softmax:
        \vspace{-1.0ex}
        \State \[
        w_j \gets 1+\frac{\exp(\alpha \cdot \Delta_j)}{\sum_{k=1}^K \exp(\alpha \cdot \Delta_k)}
        \]
        \vspace{-1.5ex}
    \EndIf
\EndFor
\end{algorithmic}
\end{algorithm}

\subsection{Majority-Guided Advantage Scaling}
While GRPO models relative preferences between completions for a given prompt, it treats all training prompts equally, regardless of their difficulty. However, not all samples offer equal learning utility. Some prompts are difficult, as evidenced by consistently poor completions; in such cases, high-quality generations are both rare and informative. Conversely, other prompts are comparatively easy, where most generations perform well, and errors on such prompts may indicate critical model failures.

To address this, we introduce a Majority-Guided Advantage Scaling \textbf{(MGAS)} mechanism, which adjusts the advantage magnitude based on the inferred difficulty of each prompt. Specifically, for each prompt group, we aggregate the predicted sub-scores from all $G$ completions and perform majority voting for each of the $K=6$ sub-score dimensions. For each sub-score dimension, we aggregate all predicted values across the group and compute the majority vote. This majority-voted value is then compared against the corresponding ground-truth label. If the majority prediction fails to match the ground truth under a defined agreement threshold, we consider it a hard case.

We average the correctness across all six aspects to compute a majority-selected score $\gamma$. This score reflects how easy the prompt is—i.e., how often the majority prediction per aspect matches the ground truth. Formally, let \( P^{(j)}=[p^{(j)}_{1},p^{(j)}_{2},...,p^{(j)}_{G}] \) denote the group of predicted values for the \( j \)-th sub-score aspects across \( G \) completions, and let \( y^{(j)} \) be the ground truth value. We define the majority-selected score \( \gamma \) as:
\begin{equation}
\gamma = \frac{1}{K} \sum_{j=1}^{K} \mathbf{1} \left[ \operatorname{mode}\left( P^{(j)} \right) = y^{(j)} \right],
\end{equation}
where \( \operatorname{mode}(\cdot) \) returns the most frequent value in the group. The score \( \gamma \in [0, 1] \) reflects the proportion of sub-score aspects where the majority prediction matches the ground truth. The scaling method is defined as:
\begin{equation}
s_i(\gamma) = \phi_{-} + (\phi_{+} - \phi_{-}) \cdot \left(1 + (\psi_i(\gamma) - c)\right)^{-\beta}.
\end{equation}

\noindent
Here, \( \phi_{-} \) and \( \phi_{+} \) are the lower and upper bounds of the scaling. \( c \) is a difficulty threshold, and \( \beta \) controls the sharpness of the modulation. The function \( \psi(\gamma) \) is defined as:
\begin{equation}
\psi_i(\gamma) =
\begin{cases}
\gamma, & \hat{A}_{i,t} > 0 \\
1 - \gamma, & \hat{A}_{i,t} < 0
\end{cases}    
\end{equation}

The final updated advantages can be defined as:
\begin{equation}
\hat{A}_{i,t}' = s_i(\gamma) \cdot \hat{A}_{i,t}.
\end{equation}

\begin{algorithm}[h]
\small
\caption{\small Majority-Guided Advantage Scaling}
\label{alg:majority_guided}
\begin{algorithmic}[0]
\State \textbf{Input:} A Group of Predicted sub-scores $P$, ground truth $y$, original advantages $\hat{A}_{i,t}$, scaling parameters $(\phi_{-}, \phi_{+}, c)$
\State \textbf{Output:} Scaled advantages $\hat{A}'_{i,t}$
\For{each aspect $j = 1,\dots,K$}
    \State Collect predictions $ P^{(j)}=[p^{(j)}_{1},p^{(j)}_{2},...,p^{(j)}_{G}]$
    \State Compute majority prediction: $m^{(j)} \gets \text{mode}(P^{(j)})$
    \State $\gamma^{(j)} \gets \mathbf{1}[m^{(j)} = y^{(j)}]$
\EndFor
\State Compute agreement: $\gamma \gets \frac{1}{K} \sum_{j=1}^K \gamma^{(j)}$
\For{each completion $i = 1,\dots,G$}
    \If{$\hat{A}_{i,t} > 0$}
        \State $\psi_i \gets \gamma$
    \Else
        \State $\psi_i \gets 1 - \gamma$
    \EndIf
    \State Compute scaling factor:
    \State $s_i \gets \phi_{-} + (\phi_{+} - \phi_{-}) \cdot \left(1 + (\psi_i(\gamma) - c)\right)^{-\beta}.$
    \State Update advantages:
    \State $\hat{A}'_{i,t} \gets  s_i \cdot \hat{A}_{i,t}$
\EndFor
\end{algorithmic}
\end{algorithm}

In summary, we introduce two strategies to enhance optimization in GRPO-based training. \textbf{Sub-score Dynamic Weighting} focuses learning on difficult error types by adjusting weights based on F1 performance. \textbf{Majority-Guided Advantage Scaling} amplifies gradients for correct completions on hard prompts and downweights errors on easy ones. Together, these mechanisms guide learning toward clinically meaningful and robust behavior.

\section{Experiments and Results}

\subsection{Datasets}
\noindent\textbf{Training Data Generation.}~~
We sample 1,000 radiology reports from the MIMIC-CXR dataset to serve as ground-truth anchors. For each case, we prompt GPT-4, previously shown to align closely with expert radiologists in evaluation tasks~\cite{liu2024mrscore}, to generate synthetic diagnostic reports exhibiting varied error profiles. We control report quality by systematically injecting clinical errors (e.g., omissions, false findings, localization mistakes) to produce reports of varying fidelity. Specifically, GPT-4 is instructed to generate:
(1) high-quality reports containing 0–1 errors;
(2) medium-quality reports containing 2–3 errors;
(3) low-quality reports containing 4 or more errors.

This error-count–based prompting enables fine-grained control over semantic fidelity, ensuring comprehensive coverage across clinically plausible quality levels. In total, we collect 3,968 labeled report completions, each paired with its corresponding ground-truth anchor. Detailed prompt is in Appendix~\ref{sec:appendix}.

\noindent\textbf{ReXVal}~\citep{yu2023radiology} is a publicly available benchmark designed to evaluate alignment between automated metrics and expert human judgments in radiology report assessment. It consists of 200 candidate–reference report pairs sampled from 50 MIMIC-CXR studies, with four candidate reports per study. Each pair is annotated by six board-certified radiologists using the six-category RadCliQ error taxonomy, which provides fine-grained sub-score annotations based on the number of errors in each category.

\subsection{Experimental Settings}
We evaluate our proposed metric against a diverse set of established baselines, including traditional NLG metrics (BLEU-4~\citep{papineni2002bleu}, ROUGE-L~\citep{lin2004rouge}, METEOR~\citep{banerjee2005meteor}), semantic similarity metrics (BERTScore~\citep{zhang2019bertscore}), and clinically informed or structure-aware metrics such as RadGraph F1~\citep{jain2021radgraph}, Semb score, and RadCliQ-v1~\citep{yu2023evaluating}. Following prior work~\citep{yu2023radiology}, we assess alignment with human judgment by computing Kendall’s Tau and Spearman correlation between metric outputs and expert annotations on the ReXVal benchmark.

We adopt Qwen2.5-7B\footnote{https://huggingface.co/Qwen/Qwen2.5-7B}~\cite{qwen2.5} as the base language model. Fine-tuning is conducted using LoRA~\cite{hu2022lora} with a rank of 16 and a scaling factor of $\alpha=16$. We set the learning rate to $1\mathrm{e}{-5}$, batch size to 4, and train for 2 epochs using 2$\times$NVIDIA RTX A6000 (48GB) GPUs. For Majority-Guided Advantage Scaling, we use $\phi_{-}=0.8$, $\phi_{+}=1.2$, and difficulty threshold $c=0.5$.

\begin{figure*}[t]
\centering
\centerline{\includegraphics[width=1\linewidth]{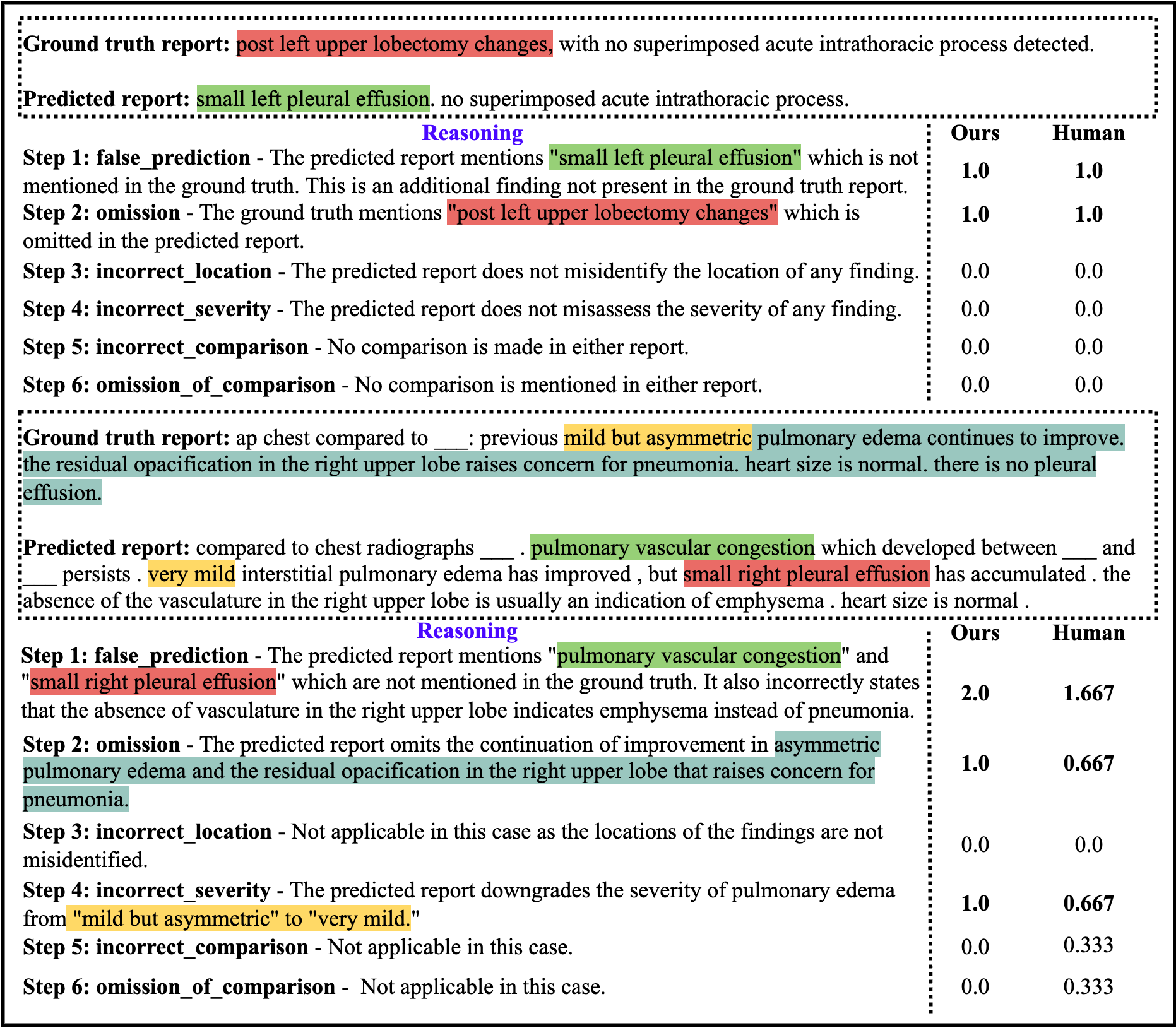}}
\caption{Case studies comparing our model’s sub-score predictions with human annotations, with step-by-step reasoning shown for each of the six clinical error types. Color highlights mark reasoning findings.}
\label{fig:case}
\end{figure*}
\subsection{Main Results}

\begin{table*}[h]
  \centering
  \resizebox{0.7\linewidth}{!}{%
  \begin{tabular}{lcc}
    \hline
    \textbf{Metric} & \textbf{Kendall’s Tau$\uparrow$} & \textbf{Spearman$\uparrow$} \\
    \hline
    BLEU-4~\citep{papineni2002bleu} & 0.345 & 0.475 \\
    ROUGE-L~\citep{lin2004rouge} & 0.491 & 0.663 \\
    METEOR~\citep{banerjee2005meteor} & 0.464 & 0.627 \\
    BertScore~\citep{zhang2019bertscore} & 0.507 & 0.677 \\
    RadGraphF1~\citep{jain2021radgraph} & 0.516 & 0.702 \\
    Semb\_score~\citep{yu2023evaluating} & 0.494 & 0.665 \\
    RadCliQ-v1~\citep{yu2023evaluating} & 0.631 & 0.816 \\
    GREEN~\citep{ostmeier2024green} & 0.640 & -- \\
    RaTEScore~\citep{zhao2024ratescore} & 0.527 & -- \\
    \textbf{Ours} & \textbf{0.730} & \textbf{0.871} \\
    \hline
    \multicolumn{3}{c}{\textit{Results below are not strictly comparable}} \\
    \multicolumn{3}{c}{\textit{ because they are using an online model (e.g., GPT4).}} \\
    \rowcolor{gray!30} RadFact~\citep{bannur2024maira} & 0.590 & -- \\
    \rowcolor{gray!30} CheXprompt~\citep{zambrano2025clinically} & 0.750 & -- \\
    \rowcolor{gray!30} FineRadScore~\citep{huang2024fineradscore} & 0.737 & -- \\
    \hline
  \end{tabular}
  }
    \caption{Human Correlation Comparison of Evaluation Metrics on ReXVal Dataset.}
  \label{tab:metrics_comparison}
\end{table*}

\noindent\textbf{Comparison with Existing Metrics.}~~Table~\ref{tab:metrics_comparison} presents a comparative evaluation of our proposed metric against a comprehensive set of baseline methods on the ReXVal dataset, using Kendall’s Tau and Spearman correlation coefficients to assess alignment with expert ratings. Our method achieves the highest overall performance among all non-LLM approaches, with a Kendall’s Tau of 0.730 and Spearman correlation of 0.871. It consistently outperforms traditional NLG metrics (BLEU-4, ROUGE-L, METEOR), semantic similarity scores (BERTScore, Semb Score), and structure-aware clinical metrics (RadGraph F1, RadCliQ-v1). Recent advanced models, such as GREEN (Kendall: 0.640) and RaTEScore (0.527), lack fine-grained interpretability and do not provide sub-scores, limiting their utility in detailed evaluation scenarios. Online methods such as CheXprompt and FineRadScore achieve comparable Kendall scores (0.750 and 0.737, respectively) but rely on commercial LLM APIs (e.g., GPT-4), which introduce concerns related to privacy, reproducibility, and operational cost.

\subsection{Ablation Study}

\begin{table*}[h]
\centering
\small
\setlength{\tabcolsep}{4pt}
\renewcommand{\arraystretch}{1.2}
\begin{tabular}{lcccccccc}
\toprule
\textbf{Criteria} & \multicolumn{2}{c}{\textbf{Baseline}} & \multicolumn{2}{c}{\textbf{+ GRPO}} & \multicolumn{2}{c}{\textbf{+ SDW}} & \multicolumn{2}{c}{\textbf{+ MGAS (Ours)}} \\
\cmidrule(lr){2-3} \cmidrule(lr){4-5} \cmidrule(lr){6-7} \cmidrule(lr){8-9}
& Kendall & Spearman & Kendall & Spearman & Kendall & Spearman & Kendall & Spearman \\
\midrule
False prediction          & 0.507 & 0.581 & 0.608 & 0.704 & 0.614 & 0.704 & \textbf{0.645} & \textbf{0.742} \\
Omission of finding       & 0.323 & 0.366 & 0.576 & 0.662 & 0.583 & 0.682 & \textbf{0.596} & \textbf{0.706} \\
Incorrect location        & 0.375 & 0.401 & 0.461 & 0.482 & \textbf{0.533} & \textbf{0.571} & 0.473 & 0.506 \\
Incorrect severity        & 0.430 & 0.460 & 0.571 & 0.614 & 0.450 & 0.482 & \textbf{0.570} & \textbf{0.611} \\
Absence of comparison     & 0.106 & 0.112 & 0.170 & 0.182 & 0.176 & 0.189 & \textbf{0.186} & \textbf{0.196} \\
Omission of comparison    & 0.160 & 0.168 & 0.194 & 0.204 & \textbf{0.317} & \textbf{0.333} & 0.238 & 0.252 \\
\midrule
\textbf{Total}            & 0.495 & 0.634 & 0.690 & 0.832 & 0.698 & 0.838 & \textbf{0.730} & \textbf{0.871} \\
\bottomrule
\end{tabular}
\caption{Ablation study of reward design. Our full model combines GRPO training with Sub-score Dynamic Weighting (SDW) and Majority-Guided Advantage Scaling (MGAS), achieving the highest correlation in bold.}
\label{tab:ablation}
\end{table*}

To evaluate the contribution of each component, we perform an ablation study summarized in Table~\ref{tab:ablation}. Beginning with a baseline model trained via supervised fine-tuning, we incrementally introduce GRPO, Sub-score Dynamic Weighting (SDW), and Majority-Guided Advantage Scaling (MGAS).

Introducing GRPO substantially improves alignment with expert annotations (Kendall: 0.495~$\rightarrow$~0.690; Spearman: 0.634~$\rightarrow$~0.832), confirming the effectiveness of preference-based learning in evaluation radiology reports. We attribute part of this gain to GRPO’s ability to better capture intermediate reasoning signals during generation. Adding SDW yields further improvements in sub-scores such as \textit{Omission of finding} and \textit{Absence of comparison}, highlighting its utility in dynamically emphasizing clinically underrepresented yet critical error types. Finally, incorporating MGAS leads to the best overall performance (Kendall: \textbf{0.730}, Spearman: \textbf{0.871}). This validates our intuition that difficult prompts deserve greater reward signal amplification when answered correctly, while easy prompts should be penalized more harshly if mistakes occur. MGAS helps stabilize updates by aligning learning signals with case difficulty.

Overall, these results demonstrate the complementary benefits of SDW and MGAS in enhancing the optimization dynamics of GRPO, achieving high correlation across diverse error categories.

\subsection{Qualitative Analysis}
Figure~\ref{fig:case} provides visual examples of how our model’s sub-score predictions align with human annotations under the RadCliQ framework. In the first case, the generated report incorrectly introduces a new finding “small left pleural effusion” and omits the key phrase “post left upper lobectomy changes” (both highlighted), resulting in scores of 1.0 for both “false prediction” and “omission of finding,” fully aligned with expert ratings. In the second case, the model hallucinates “pulmonary vascular congestion” and “small right pleural effusion,” while omitting the clinically important improvement in “pulmonary edema” and “right upper lobe opacification.” These errors lead to elevated scores for “false prediction,” “omission,” and “incorrect severity,” closely mirroring human assessments. These examples illustrate the model’s ability to perform fine-grained error identification consistent with expert judgment.

\section{Conclusion}
We introduce \textbf{RadReason}, an interpretable evaluation framework for radiology reports that produces structured sub-scores and explicit reasoning across clinically meaningful error categories. By embedding \textit{Sub-score Dynamic Weighting} and \textit{Majority-Guided Advantage Scaling}, our method adaptively focuses on harder sub-aspects and calibrates learning based on prompt difficulty. Empirical results on the ReXVal benchmark demonstrate that RadReason not only surpasses prior metrics. 


\section*{Limitations}
Our evaluation is conducted on ReXVal~\cite{yu2023radiology}, the standard benchmark for radiology report assessment aligned with human judgment. While moderate in size due to the cost of expert annotation, it enables fair and meaningful comparisons across methods. We adopt the six clinically grounded error categories defined in RadCliQ~\cite{yu2023evaluating}; though fixed, our framework is modular and readily extensible to alternative or hierarchical taxonomies. While our experiments focus on chest X-ray reports from MIMIC-CXR, the reward-based approach is modality-agnostic and can generalize to structured diagnostic outputs from CT, MRI, or multimodal reports. Looking forward, the sub-score-based reward formulation may also inspire evaluation methods for other clinical generation tasks such as medical VQA.

\bibliography{new_version}

\begin{thebibliography}{30}
\providecommand{\natexlab}[1]{#1}

\bibitem[{Achiam et~al.(2023)Achiam, Adler, Agarwal, Ahmad, Akkaya, Aleman, Almeida, Altenschmidt, Altman, Anadkat et~al.}]{achiam2023gpt}
Josh Achiam, Steven Adler, Sandhini Agarwal, Lama Ahmad, Ilge Akkaya, Florencia~Leoni Aleman, Diogo Almeida, Janko Altenschmidt, Sam Altman, Shyamal Anadkat, and 1 others. 2023.
\newblock Gpt-4 technical report.
\newblock \emph{arXiv preprint arXiv:2303.08774}.

\bibitem[{Banerjee and Lavie(2005)}]{banerjee2005meteor}
Satanjeev Banerjee and Alon Lavie. 2005.
\newblock Meteor: An automatic metric for mt evaluation with improved correlation with human judgments.
\newblock In \emph{Proceedings of the acl workshop on intrinsic and extrinsic evaluation measures for machine translation and/or summarization}, pages 65--72.

\bibitem[{Bannur et~al.(2024)Bannur, Bouzid, Castro, Schwaighofer, Thieme, Bond-Taylor, Ilse, P{\'e}rez-Garc{\'\i}a, Salvatelli, Sharma et~al.}]{bannur2024maira}
Shruthi Bannur, Kenza Bouzid, Daniel~C Castro, Anton Schwaighofer, Anja Thieme, Sam Bond-Taylor, Maximilian Ilse, Fernando P{\'e}rez-Garc{\'\i}a, Valentina Salvatelli, Harshita Sharma, and 1 others. 2024.
\newblock Maira-2: Grounded radiology report generation.
\newblock \emph{arXiv preprint arXiv:2406.04449}.

\bibitem[{Besta et~al.(2024)Besta, Blach, Kubicek, Gerstenberger, Podstawski, Gianinazzi, Gajda, Lehmann, Niewiadomski, Nyczyk et~al.}]{besta2024graph}
Maciej Besta, Nils Blach, Ales Kubicek, Robert Gerstenberger, Michal Podstawski, Lukas Gianinazzi, Joanna Gajda, Tomasz Lehmann, Hubert Niewiadomski, Piotr Nyczyk, and 1 others. 2024.
\newblock Graph of thoughts: Solving elaborate problems with large language models.
\newblock In \emph{Proceedings of the AAAI Conference on Artificial Intelligence}, volume~38, pages 17682--17690.

\bibitem[{Grattafiori et~al.(2024)Grattafiori, Dubey, Jauhri, Pandey, Kadian, Al-Dahle, Letman, Mathur, Schelten, Vaughan et~al.}]{grattafiori2024llama}
Aaron Grattafiori, Abhimanyu Dubey, Abhinav Jauhri, Abhinav Pandey, Abhishek Kadian, Ahmad Al-Dahle, Aiesha Letman, Akhil Mathur, Alan Schelten, Alex Vaughan, and 1 others. 2024.
\newblock The llama 3 herd of models.
\newblock \emph{arXiv preprint arXiv:2407.21783}.

\bibitem[{Guo et~al.(2025)Guo, Yang, Zhang, Song, Zhang, Xu, Zhu, Ma, Wang, Bi et~al.}]{guo2025deepseek}
Daya Guo, Dejian Yang, Haowei Zhang, Junxiao Song, Ruoyu Zhang, Runxin Xu, Qihao Zhu, Shirong Ma, Peiyi Wang, Xiao Bi, and 1 others. 2025.
\newblock Deepseek-r1: Incentivizing reasoning capability in llms via reinforcement learning.
\newblock \emph{arXiv preprint arXiv:2501.12948}.

\bibitem[{Hu et~al.(2022)Hu, Shen, Wallis, Allen-Zhu, Li, Wang, Wang, Chen et~al.}]{hu2022lora}
Edward~J Hu, Yelong Shen, Phillip Wallis, Zeyuan Allen-Zhu, Yuanzhi Li, Shean Wang, Lu~Wang, Weizhu Chen, and 1 others. 2022.
\newblock Lora: Low-rank adaptation of large language models.
\newblock \emph{ICLR}, 1(2):3.

\bibitem[{Huang et~al.(2024)Huang, Banerjee, Wu, Reis, and Rajpurkar}]{huang2024fineradscore}
Alyssa Huang, Oishi Banerjee, Kay Wu, Eduardo~Pontes Reis, and Pranav Rajpurkar. 2024.
\newblock Fineradscore: A radiology report line-by-line evaluation technique generating corrections with severity scores.
\newblock In \emph{Machine Learning for Healthcare Conference}. PMLR.

\bibitem[{Huang et~al.(2023)Huang, Zhang, and Zhang}]{huang2023kiut}
Zhongzhen Huang, Xiaofan Zhang, and Shaoting Zhang. 2023.
\newblock Kiut: Knowledge-injected u-transformer for radiology report generation.
\newblock In \emph{Proceedings of the IEEE/CVF conference on computer vision and pattern recognition}, pages 19809--19818.

\bibitem[{Jain et~al.(2021)Jain, Agrawal, Saporta, Truong, Duong, Bui, Chambon, Zhang, Lungren, Ng et~al.}]{jain2021radgraph}
Saahil Jain, Ashwin Agrawal, Adriel Saporta, Steven~QH Truong, Du~Nguyen Duong, Tan Bui, Pierre Chambon, Yuhao Zhang, Matthew~P Lungren, Andrew~Y Ng, and 1 others. 2021.
\newblock Radgraph: Extracting clinical entities and relations from radiology reports.
\newblock \emph{arXiv preprint arXiv:2106.14463}.

\bibitem[{Kumar et~al.(2025)Kumar, Ashraf, Thawakar, Anwer, Cholakkal, Shah, Yang, Torr, Khan, and Khan}]{kumar2025llm}
Komal Kumar, Tajamul Ashraf, Omkar Thawakar, Rao~Muhammad Anwer, Hisham Cholakkal, Mubarak Shah, Ming-Hsuan Yang, Phillip~HS Torr, Fahad~Shahbaz Khan, and Salman Khan. 2025.
\newblock Llm post-training: A deep dive into reasoning large language models.
\newblock \emph{arXiv preprint arXiv:2502.21321}.

\bibitem[{Li et~al.(2023)Li, Liu, Wang, Liang, Liu, Wang, Cui, Tu, Wang, and Zhou}]{li2023comprehensive}
Yingshu Li, Yunyi Liu, Zhanyu Wang, Xinyu Liang, Lingqiao Liu, Lei Wang, Leyang Cui, Zhaopeng Tu, Longyue Wang, and Luping Zhou. 2023.
\newblock A comprehensive study of gpt-4v’s multimodal capabilities in medical imaging.
\newblock \emph{medRxiv}, pages 2023--11.

\bibitem[{Li et~al.(2024)Li, Wang, Liu, Wang, Liu, and Zhou}]{li2024kargen}
Yingshu Li, Zhanyu Wang, Yunyi Liu, Lei Wang, Lingqiao Liu, and Luping Zhou. 2024.
\newblock Kargen: Knowledge-enhanced automated radiology report generation using large language models.
\newblock In \emph{International Conference on Medical Image Computing and Computer-Assisted Intervention}, pages 382--392. Springer.

\bibitem[{Lin(2004)}]{lin2004rouge}
Chin-Yew Lin. 2004.
\newblock Rouge: A package for automatic evaluation of summaries.
\newblock In \emph{Text summarization branches out}, pages 74--81.

\bibitem[{Liu et~al.(2024)Liu, Wang, Li, Liang, Liu, Wang, and Zhou}]{liu2024mrscore}
Yunyi Liu, Zhanyu Wang, Yingshu Li, Xinyu Liang, Lingqiao Liu, Lei Wang, and Luping Zhou. 2024.
\newblock Mrscore: Evaluating medical report with llm-based reward system.
\newblock In \emph{International Conference on Medical Image Computing and Computer-Assisted Intervention}, pages 283--292. Springer.

\bibitem[{Ostmeier et~al.(2024)Ostmeier, Xu, Chen, Varma, Blankemeier, Bluethgen, Md, Moseley, Langlotz, Chaudhari et~al.}]{ostmeier2024green}
Sophie Ostmeier, Justin Xu, Zhihong Chen, Maya Varma, Louis Blankemeier, Christian Bluethgen, Arne Md, Michael Moseley, Curtis Langlotz, Akshay Chaudhari, and 1 others. 2024.
\newblock Green: Generative radiology report evaluation and error notation.
\newblock In \emph{Findings of the Association for Computational Linguistics: EMNLP 2024}, pages 374--390.

\bibitem[{Papineni et~al.(2002)Papineni, Roukos, Ward, and Zhu}]{papineni2002bleu}
Kishore Papineni, Salim Roukos, Todd Ward, and Wei-Jing Zhu. 2002.
\newblock Bleu: a method for automatic evaluation of machine translation.
\newblock In \emph{Proceedings of the 40th annual meeting of the Association for Computational Linguistics}, pages 311--318.

\bibitem[{Rafailov et~al.(2023)Rafailov, Sharma, Mitchell, Manning, Ermon, and Finn}]{rafailov2023direct}
Rafael Rafailov, Archit Sharma, Eric Mitchell, Christopher~D Manning, Stefano Ermon, and Chelsea Finn. 2023.
\newblock Direct preference optimization: Your language model is secretly a reward model.
\newblock \emph{Advances in Neural Information Processing Systems}, 36:53728--53741.

\bibitem[{Shao et~al.(2024)Shao, Wang, Zhu, Xu, Song, Bi, Zhang, Zhang, Li, Wu et~al.}]{shao2024deepseekmath}
Zhihong Shao, Peiyi Wang, Qihao Zhu, Runxin Xu, Junxiao Song, Xiao Bi, Haowei Zhang, Mingchuan Zhang, YK~Li, Y~Wu, and 1 others. 2024.
\newblock Deepseekmath: Pushing the limits of mathematical reasoning in open language models.
\newblock \emph{arXiv preprint arXiv:2402.03300}.

\bibitem[{Smit et~al.(2020)Smit, Jain, Rajpurkar, Pareek, Ng, and Lungren}]{smit2020chexbert}
Akshay Smit, Saahil Jain, Pranav Rajpurkar, Anuj Pareek, Andrew~Y Ng, and Matthew~P Lungren. 2020.
\newblock Chexbert: combining automatic labelers and expert annotations for accurate radiology report labeling using bert.
\newblock \emph{arXiv preprint arXiv:2004.09167}.

\bibitem[{Team(2024)}]{qwen2.5}
Qwen Team. 2024.
\newblock \href {https://qwenlm.github.io/blog/qwen2.5/} {Qwen2.5: A party of foundation models}.

\bibitem[{Wang et~al.(2024)Wang, Wang, Li, Ma, Wang, Jiang, Li, and Tang}]{wang2024cxpmrg}
Xiao Wang, Fuling Wang, Yuehang Li, Qingchuan Ma, Shiao Wang, Bo~Jiang, Chuanfu Li, and Jin Tang. 2024.
\newblock Cxpmrg-bench: Pre-training and benchmarking for x-ray medical report generation on chexpert plus dataset.
\newblock \emph{arXiv preprint arXiv:2410.00379}.

\bibitem[{Wei et~al.(2022)Wei, Wang, Schuurmans, Bosma, Xia, Chi, Le, Zhou et~al.}]{wei2022chain}
Jason Wei, Xuezhi Wang, Dale Schuurmans, Maarten Bosma, Fei Xia, Ed~Chi, Quoc~V Le, Denny Zhou, and 1 others. 2022.
\newblock Chain-of-thought prompting elicits reasoning in large language models.
\newblock \emph{Advances in neural information processing systems}, 35:24824--24837.

\bibitem[{Yang et~al.(2024)Yang, Yang, Zhang, Hui, Zheng, Yu, Li, Liu, Huang, Wei et~al.}]{yang2024qwen2}
An~Yang, Baosong Yang, Beichen Zhang, Binyuan Hui, Bo~Zheng, Bowen Yu, Chengyuan Li, Dayiheng Liu, Fei Huang, Haoran Wei, and 1 others. 2024.
\newblock Qwen2. 5 technical report.
\newblock \emph{arXiv preprint arXiv:2412.15115}.

\bibitem[{Yao et~al.(2023)Yao, Yu, Zhao, Shafran, Griffiths, Cao, and Narasimhan}]{yao2023tree}
Shunyu Yao, Dian Yu, Jeffrey Zhao, Izhak Shafran, Tom Griffiths, Yuan Cao, and Karthik Narasimhan. 2023.
\newblock Tree of thoughts: Deliberate problem solving with large language models.
\newblock \emph{Advances in neural information processing systems}, 36:11809--11822.

\bibitem[{Yu et~al.(2023{\natexlab{a}})Yu, Endo, Krishnan, Pan, Tsai, Reis, Fonseca, Lee, Abad, Ng et~al.}]{yu2023evaluating}
Feiyang Yu, Mark Endo, Rayan Krishnan, Ian Pan, Andy Tsai, Eduardo~Pontes Reis, Eduardo Kaiser Ururahy~Nunes Fonseca, Henrique Min~Ho Lee, Zahra Shakeri~Hossein Abad, Andrew~Y Ng, and 1 others. 2023{\natexlab{a}}.
\newblock Evaluating progress in automatic chest x-ray radiology report generation.
\newblock \emph{Patterns}, 4(9).

\bibitem[{Yu et~al.(2023{\natexlab{b}})Yu, Endo, Krishnan, Pan, Tsai, Reis, Fonseca, Lee, Shakeri, Ng et~al.}]{yu2023radiology}
Feiyang Yu, Mark Endo, Rayan Krishnan, Ian Pan, Andy Tsai, Eduardo~Pontes Reis, EKU Fonseca, Henrique Lee, Zahra Shakeri, Andrew Ng, and 1 others. 2023{\natexlab{b}}.
\newblock Radiology report expert evaluation (rexval) dataset.

\bibitem[{Zambrano~Chaves et~al.(2025)Zambrano~Chaves, Huang, Xu, Xu, Usuyama, Zhang, Wang, Xie, Khademi, Yang et~al.}]{zambrano2025clinically}
Juan~Manuel Zambrano~Chaves, Shih-Cheng Huang, Yanbo Xu, Hanwen Xu, Naoto Usuyama, Sheng Zhang, Fei Wang, Yujia Xie, Mahmoud Khademi, Ziyi Yang, and 1 others. 2025.
\newblock A clinically accessible small multimodal radiology model and evaluation metric for chest x-ray findings.
\newblock \emph{Nature Communications}, 16(1):3108.

\bibitem[{Zhang et~al.(2019)Zhang, Kishore, Wu, Weinberger, and Artzi}]{zhang2019bertscore}
Tianyi Zhang, Varsha Kishore, Felix Wu, Kilian~Q Weinberger, and Yoav Artzi. 2019.
\newblock Bertscore: Evaluating text generation with bert.
\newblock \emph{arXiv preprint arXiv:1904.09675}.

\bibitem[{Zhao et~al.(2024)Zhao, Wu, Zhang, Zhang, Wang, and Xie}]{zhao2024ratescore}
Weike Zhao, Chaoyi Wu, Xiaoman Zhang, Ya~Zhang, Yanfeng Wang, and Weidi Xie. 2024.
\newblock Ratescore: A metric for radiology report generation.
\newblock In \emph{Proceedings of the 2024 Conference on Empirical Methods in Natural Language Processing}, pages 15004--15019.

\end{thebibliography}

\newpage
\appendix
\section{Appendix}
\label{sec:appendix}
\subsection{GPT-4 Prompt Template}
\begin{figure*}[!htbp]
\centering
\includegraphics[width=\linewidth]{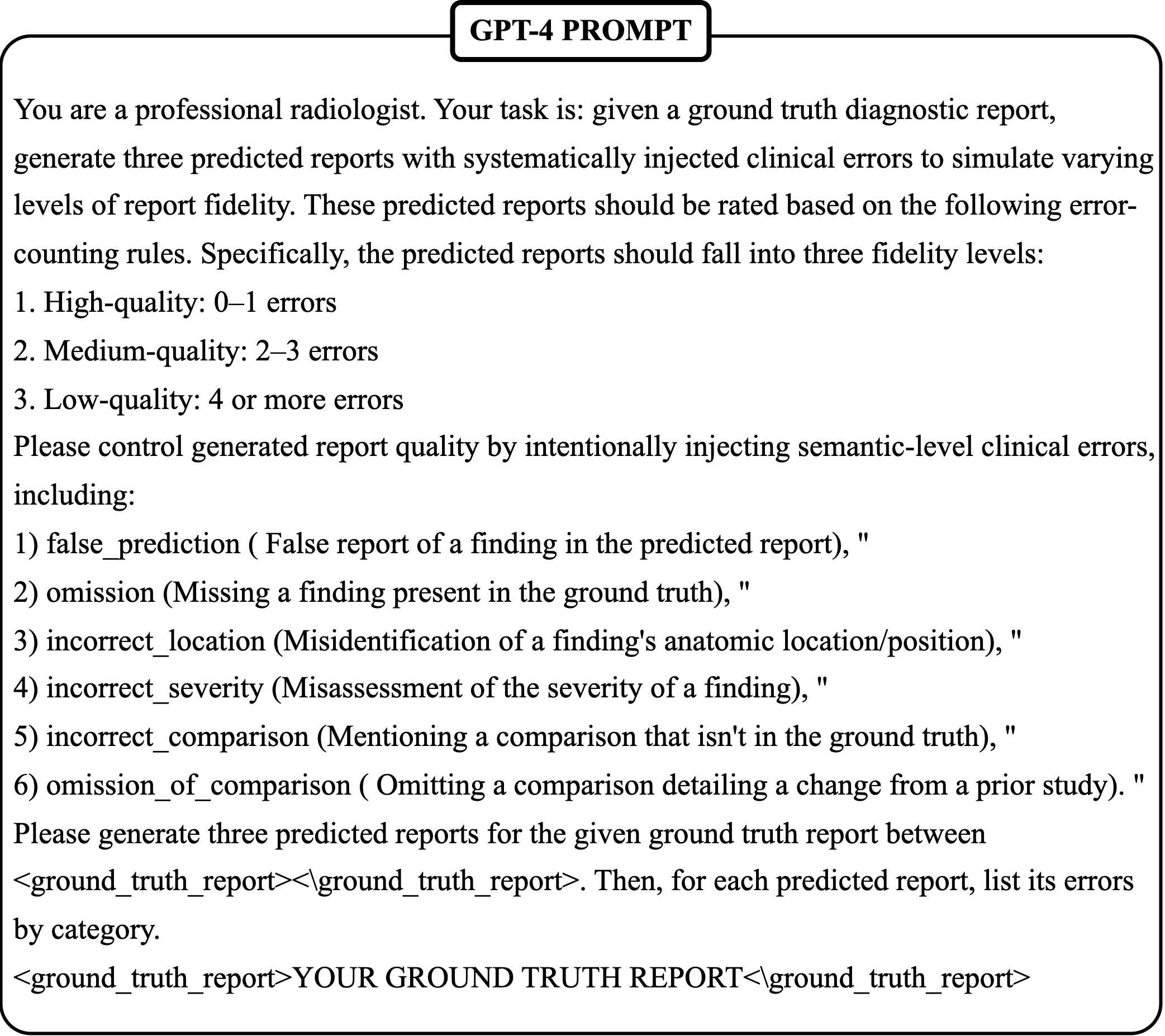}
\caption{GPT-4 Prompt Example.}
\label{fig:appdx}
\end{figure*}

The prompt in Figure~\ref{fig:appdx} was used in GPT-4 to generate the training data.

\end{document}